\documentclass{article}
\usepackage{spconf,amsmath,graphicx,hyperref}

\usepackage{bbding}
\usepackage{bbm}
\usepackage{graphicx}
\usepackage{subcaption}
\usepackage{caption}
\usepackage{xcolor}
\usepackage{amsfonts}
\usepackage{booktabs}

\definecolor{blush}{rgb}{0.87, 0.36, 0.51}
\newcommand\best[1]{\textbf{\color{blush}#1}}

\title{Learning Fair Domain Adaptation with Virtual Label Distribution}

\name{Yuguang Zhang$^{1,2,4}$\thanks{Email: zhangyuguang2020@ia.ac.cn}, Lijun Sheng$^{2,3}$, Jian Liang$^{1,2}$\thanks{Jian Liang is the corresponding author.}, Ran He$^{1,2}$}
  
\address{
$^{1}$ School of Artificial Intelligence, University of Chinese Academy of Sciences \\
$^{2}$ NLPR \& MAIS, Institute of Automation, Chinese Academy of Sciences \\
$^{3}$ University of Science and Technology of China \\
$^{4}$ China Electronics Standardization Insitute \\
}

\begin{document}
%\ninept
%
\maketitle
\begin{abstract}
Unsupervised Domain Adaptation (UDA) aims to mitigate performance degradation when training and testing data are sampled from different distributions.
While significant progress has been made in enhancing overall accuracy, most existing methods overlook performance disparities across categories—an issue we refer to as category fairness.
Our empirical analysis reveals that UDA classifiers tend to favor certain easy categories while neglecting difficult ones.
To address this, we propose Virtual Label-distribution-aware Learning (VILL), a simple yet effective framework designed to improve worst-case performance while preserving high overall accuracy.
The core of VILL is an adaptive re-weighting strategy that amplifies the influence of hard-to-classify categories.
Furthermore, we introduce a KL-divergence-based re-balancing strategy, which explicitly adjusts decision boundaries to enhance category fairness.
Experiments on commonly used datasets demonstrate that VILL can be seamlessly integrated as a plug-and-play module into existing UDA methods, significantly improving category fairness.
\end{abstract}

\section{Introduction}
\label{sec:intro}

Deep neural networks \cite{he2016deep, radford2021learning} have achieved remarkable success in various visual tasks, but always suffer from performance degradation by the distribution gap between training and deployment domains \cite{ben2010theory, quinonero2022dataset}.
To enhance generalization and avoid the heavy annotation cost, researchers propose Unsupervised Domain Adaptation (UDA) \cite{singhal2023domain, ganin2016domain} to utilize labeled data from a source domain to learn tasks in an unlabeled target domain.
Numerous UDA approaches have been developed, yielding promising results across a range of visual tasks, including image classification \cite{tzeng2017adversarial, cicek2019unsupervised, chang2019domain}, semantic segmentation \cite{zou2018unsupervised,vu2019advent}, object detection \cite{kim2019self, vs2021mega}, and person re-identification \cite{ge2019mutual}. 

\begin{figure}[t]
	\centering
	\includegraphics[width=0.45\textwidth]{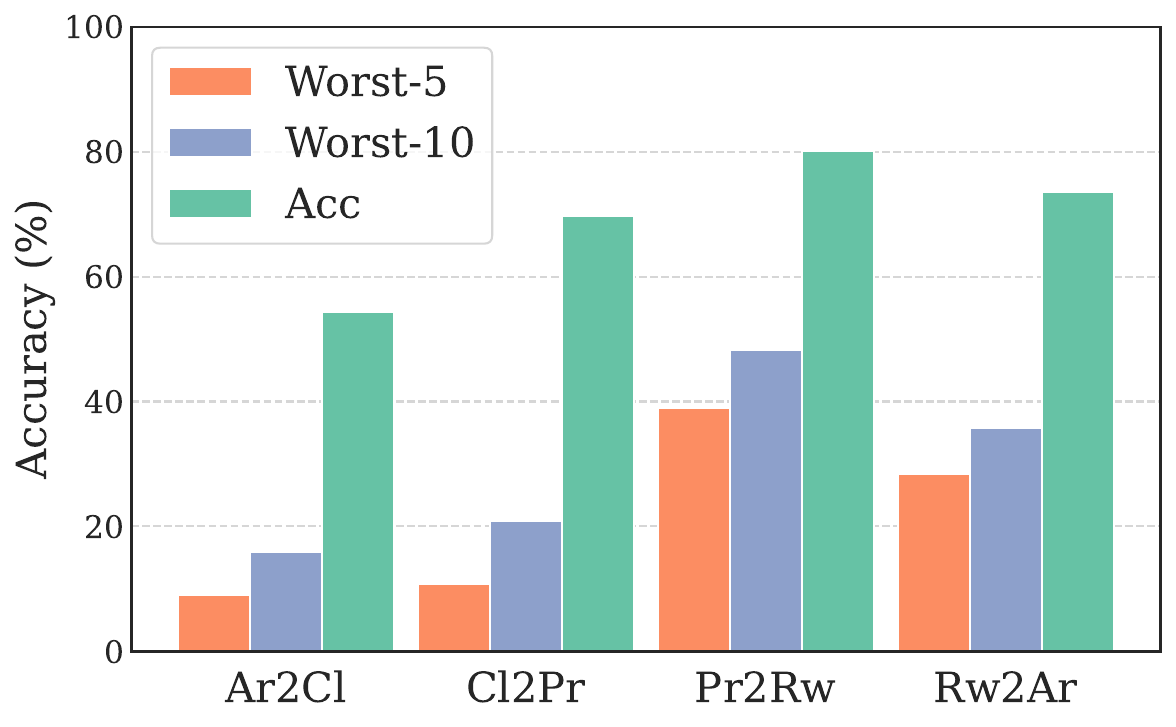}
	\caption{Results of WAcc-5, WAcc-10 and conventional accuracy of CDAN \cite{long2018conditional} on 4 tasks from OfficeHome dataset. Worst-$N$ is defined as the average accuracy of $N$ classes with the worst accuracy in the target domain.}
	\label{fig:cdan intro}
	\vspace{-0.5cm}
\end{figure}

In current research, UDA algorithms are typically evaluated based on the average target classification accuracy.
This simple metric conceals significant disparities in category performance—a problem we refer to as poor category fairness.
To better assess this issue, we introduce a more sensitive metric: the worst-N accuracy, defined as the average accuracy of the $N$ lowest-performing classes.
For example, consider the CDAN algorithm \cite{long2018conditional}, which achieves an average accuracy of 68.0\% across 12 transfer tasks on the OfficeHome dataset \cite{venkateswara2017deep}.
However, its worst-5 accuracy is only 20.3\%, revealing a substantial performance gap (we provide results of four tasks in Figure~\ref{fig:cdan intro}).
This discrepancy raises critical concerns.
In tasks where category divisions align with sensitive attributes, poor category fairness may lead to ethical issues related to bias and discrimination \cite{dwork2012fairness}.
Moreover, in safety-critical applications, worst-case performance often dictates system reliability—a phenomenon known as the Buckets Effect.
For instance, a segmentation model with high average accuracy may still be unacceptable if it consistently misclassifies crucial categories such as Person.

To address poor category fairness during the adaptation process, we propose a simple plug-and-play method called Virtual Label-distribution-aware Learning (VILL).
Inspired by re-weighting strategies from long-tail learning \cite{cui2019class}, we introduce a virtual label-distribution-aware re-weighting mechanism to mitigate category bias.
Specifically, we assign high weights to potentially underperforming categories to strengthen their representation.
These weights are derived from a virtual label distribution, which is calculated from the category distribution of target pseudo-labels.
The virtual label distribution is updated at the start of each training epoch, allowing the weight vector to dynamically incorporate the latest target domain statistics.

Furthermore, we propose a label-distribution-aware re-balancing strategy to explicitly refine the decision boundaries of under-represented categories.
Specifically, we measure the disparity between the target model’s output and the category weight vector using the Kullback–Leibler (KL) divergence.
Minimizing this divergence encourages the network to generate more diverse predictions and shift its focus toward the virtual minority categories.
Since the re-balancing strategy does not rely on labeled target data, it enables explicit optimization of decision boundaries for under-represented categories, thereby enhancing category-level fairness without the need for additional supervision.

Our key contributions are summarized as follows:
\begin{itemize}
\item We conduct the first systematic investigation of category fairness in UDA and propose worst-case accuracy as a new evaluation metric.
\item We introduce VILL, a generic plug-and-play framework for fair domain adaptation, consisting of a re-weighting mechanism and a re-balancing strategy.
\item Extensive results demonstrate that integrating VILL into UDA methods leads to better category fairness.
\end{itemize}

\section{Methodology}
\subsection{Preliminaries}
In the typical UDA setting, we are given access to a source domain $\mathcal{D}_s = \{(x_i^s, y_i^s)\}_{i=1}^{N_s}$ with labeled samples, and a target domain $\mathcal{D}_t = \{x_i^t\}_{i=1}^{N_t}$ containing only unlabeled data.
We focus on the standard closed-set UDA scenario, where both domains share the same label space consisting of $C$ categories.
Most UDA approaches adopt a network architecture composed of a feature extractor $G: \mathcal{X}_s\to \mathbb{R}^{d}$ and a classifier $F: \mathbb{R}^{d}\to \mathbb{R}^{C}$ \cite{ganin2016domain}. 
Given an input sample $x_i$, the extracted feature is denoted by $z_i=G(x_i)$, the classifier's output logits as $q_i=F(G(x_i))$ , and the predicted class probabilities as $p_i={\sigma}(q_i)$, where $\sigma$ denotes the softmax function.

Most UDA algorithms minimize the empirical source risk and design a few adaptation objectives to align two domains \cite{ganin2016domain, long2018conditional}.
This is typically achieved by optimizing a combined loss function:
\begin{equation}
    \mathcal{L} = \mathcal{L}_{CE}(\mathcal{D}_s) + \mathcal{L}_{DA}(\mathcal{D}_s, \mathcal{D}_t),
    \label{eq:uda basic}
\end{equation}
where $\mathcal{L}_{CE}$ is the cross-entropy loss computed on the source domain, and $\mathcal{L}_{DA}$ represents the domain alignment term, commonly implemented via domain adversarial training \cite{ganin2016domain, long2018conditional} or discrepancy-based measures \cite{zhang2019bridging}.

While existing approaches have made significant strides in improving overall target performance, they often overlook imbalances across categories within the target domain.
To address this limitation, we propose \textbf{VI}rtual \textbf{L}abel-distribution-aware \textbf{L}earning (VILL), a unified framework designed to improve category fairness in UDA.

\subsection{Re-weighting Mechanism for Fair DA}
Re-weighting is a widely adopted technique for mitigating model bias toward specific categories.
The central challenge lies in designing an effective category weight vector that adjusts the learning dynamics without introducing instability.

In order to understand the overall prediction distribution of the target domain, We define the virtual label distribution as the empirical class distribution computed from pseudo labels.
Formally, let $\hat{y}_j^t$ denote the pseudo label assigned to the $j$-th target sample by the current model.
The virtual label distribution ${v} \in \mathbb{R}^C$ over $C$ categories is then given by:
\begin{equation}
v_i = \frac{N_i}{N_t}, \ \ N_i=\sum\nolimits_{j=1}^{N_t} \mathbbm{1}(\hat{y}_j^t = i), \ \ i=1,...,C,
\end{equation}
where $N_t$ is the total number of target samples.
A straightforward approach is to use the reciprocal of the virtual label distribution directly.
However, this naive method preserves magnitude disparities across categories, potentially leading to severe training instability.

To address this, we introduce a scaling strategy that normalizes the virtual label distribution relative to the mean category frequency.
Specifically, we define the scaled distribution as:
\begin{equation}
    {E_i} = \frac{N_i}{N_t}{C}, \; i=1,\dots,C,
    \label{eq:scaled distribution}
\end{equation}
where $E_i$ denotes the scaled virtual label distribution, which is the ratio between the sample count for class $i$ and the mean sample count across all categories.
To emphasize minority classes, we apply a normalized negative exponential transformation to construct the category weight vector $\boldsymbol{\omega} \in \mathbb{R}^C$:
\begin{equation}
    {\omega_i} = \frac{{1+\alpha e^{-E_i}}}{\sum\nolimits_{k=1}^{C}{(1+\alpha e^{-E_k}})}, \;  i=1,\dots,C,
    \label{eq:weight}
\end{equation}
where $\alpha$ is a hyperparameter controlling the degree of re-weighting disparity. We set $\alpha = 5.0$ by default.

Since ground-truth labels are only available in the source domain, we apply the re-weighting mechanism during source training,  encouraging a more balanced learning trajectory.
The re-weighted cross-entropy loss $\mathcal{L}_{RW}$ is defined as:
\begin{equation}
    \mathcal{L}_{RW}(x^s, y^s, \omega) = \frac{\sum\nolimits_{i=1}^{B}{\omega_{y_i^s}{{l}_{ce}{(x_i^s,y_i^s)}}}}{\sum\nolimits_{i=1}^{B}{{\omega_{y_i^s}}}},
    \label{eq:ce}
\end{equation}
where $B$ refers to the batch size, and ${l}_{ce}$ is the standard cross-entropy loss for individual samples.

Unlike conventional long-tailed learning strategies that rely on fixed class distributions known a priori, our method dynamically updates the virtual label distribution and corresponding weights after each training epoch.
This dynamic adjustment allows the model to adapt to dynamic pseudo-label distributions in the target domain, maintaining a form of self-regulating fairness.
We assign a uniform distribution to all categories as the initialization at the beginning of training.

\subsection{Re-balancing Strategy for Fair DA}
While the re-weighting mechanism effectively enhances minority categories, its impact on the target domain remains limited.
During training, the model quickly learns to predict labeled source data accurately, resulting in a weak supervised signal.
Consequently, this loss gradually loses its effectiveness in correcting category bias.

\setlength{\tabcolsep}{3.0pt}
	\begin{table*}[!ht]
    	\caption{Worst-5 and worst-10 accuracy (\%) on OfficeHome (ResNet-50). The upper table shows the worst-5 results, and the lower table shows the worst-10 results. WAcc-5 is short for the worst-5 accuracy.}
		\label{table:home-worst}
		\small
		\centering
		\vspace{-3pt}
		\resizebox{0.9\textwidth}{!}{
			\begin{tabular}{l|c| cccccccccccc | cc}
			\hline
            WAcc-5 & Backbone &{A$\to$C} &{A$\to$P} &{A$\to$R} &{C$\to$A} &{C$\to$P} &{C$\to$R} &{P$\to$A} &{P$\to$C} &{P$\to$R} &{R$\to$A} &{R$\to$C} &{R$\to$P} &{Avg.} & {} \\
			\hline
			CDAN & ResNet-50 & 9.0  & 15.0 & 34.6 & 13.0 & 10.8 & 16.1 & 11.5 & 3.5  & 39.0 & 28.4 & 21.8 & 41.3 & 20.3 \\
                + VILL & ResNet-50  & 12.0 & 28.7 & 44.7 & 15.6 & 23.0 & 27.1 & 17.5 & 6.5  & 40.3 & 38.7 & 24.6 & 43.1 & \best{26.8} \\
                \hline
                MDD & ResNet-50     & 7.5  & 16.3 & 35.2 & 10.0 & 4.9  & 19.7 & 15.7 & 5.0  & 37.0 & 30.9 & 23.4 & 40.9 & 20.5 \\
                + VILL & ResNet-50  & 10.5 & 20.9 & 43.8 & 10.2 & 23.7 & 24.0 & 18.1 & 10.0 & 37.2 & 33.4 & 24.2 & 44.1 & \best{25.0} \\
                \hline
                ATDOC & ResNet-50   & 14.3 & 17.9 & 39.1 & 4.1  & 15.3 & 28.0 & 2.4  & 11.2 & 39.3 & 17.4 & 13.6 & 41.1 & 20.3 \\
                + VILL & ResNet-50  & 12.9 & 22.4 & 41.2 & 9.1  & 33.4 & 33.0 & 3.9  & 11.5 & 41.1 & 18.4 & 16.9 & 43.2 & \best{23.9} \\
                \hline
                PDA & CLIP-ResNet-50 & 9.9  & 36.6 & 48.9 & 24.3 & 40.0 & 45.3 & 26.0 & 12.2 & 55.0 & 34.7 & 14.3 & 46.9 & 32.8 \\
                + VILL & CLIP-ResNet-50 & 17.1 & 38.8 & 45.4 & 26.7 & 33.5 & 50.4 & 28.8 & 20.8 & 56.1 & 35.9 & 17.5 & 50.5 & \best{35.1} \\
                \hline
                \hline

            WAcc-10 & Backbone &{A$\to$C} &{A$\to$P} &{A$\to$R} &{C$\to$A} &{C$\to$P} &{C$\to$R} &{P$\to$A} &{P$\to$C} &{P$\to$R} &{R$\to$A} &{R$\to$C} &{R$\to$P} &{Avg.} & {Acc Avg.} \\
            \hline

            CDAN & ResNet-50    & 15.9 & 25.2 & 42.2 & 21.6 & 20.9 & 29.8 & 18.7 & 9.7  & 48.2 & 35.7 & 27.5 & 48.9 & 28.7 & \best{68.0} \\
            + VILL  & ResNet-50 & 19.9 & 36.8 & 49.2 & 25.9 & 34.1 & 36.4 & 25.7 & 13.8 & 49.8 & 44.1 & 29.8 & 49.0 & \best{34.5} & 67.9\\
            \hline
            MDD  & ResNet-50    & 17.4 & 30.7 & 45.6 & 16.0 & 20.6 & 31.1 & 23.8 & 12.0 & 46.1 & 37.1 & 28.6 & 48.4 & 29.8 & \best{68.9} \\
            + VILL  & ResNet-50 & 18.3 & 33.7 & 51.3 & 15.9 & 31.1 & 35.6 & 26.2 & 14.3 & 46.3 & 40.0 & 29.6 & 50.3 & \best{32.7} & 67.8 \\
            \hline
            ATDOC  & ResNet-50  & 23.6 & 29.6 & 49.7 & 14.4 & 29.6 & 40.9 & 11.1 & 17.0 & 50.2 & 25.3 & 22.2 & 51.9 & 30.5 & \best{72.2} \\
            + VILL  & ResNet-50 & 22.4 & 33.3 & 50.8 & 17.9 & 38.5 & 44.2 & 13.7 & 17.1 & 51.0 & 26.7 & 23.3 & 53.2 & \best{32.7} & \best{72.2} \\
            \hline
            PDA  & CLIP-ResNet-50    & 15.7 & 44.8 & 57.0 & 33.4 & 49.9 & 53.3 & 34.0 & 17.9 & 59.3 & 38.6 & 18.7 & 54.7 & 39.8 & 74.5 \\
            + VILL & CLIP-ResNet-50 & 22.9 & 46.4 & 54.9 & 34.3 & 44.6 & 55.7 & 35.6 & 24.0 & 59.8 & 39.8 & 22.4 & 57.4 & \best{41.5} & \best{75.0} \\
            \hline
            
            \end{tabular}
		}
	\end{table*}

To address this limitation, we introduce a more explicit optimization objective for the target domain.
Building upon the re-weighting mechanism, we propose a KL-divergence-based loss that explicitly adjusts the decision boundary by aligning the model's predictions with an ideal target distribution:
\begin{equation}
\begin{split}
D_{KL}(\overline{p} \| \omega)&= - \sum\limits_{k=1}^{C}{\overline{p}_{k}log\omega_{k}}+\sum\limits_{k=1}^{C}{\overline{p}_{k}log\overline{p}_{k}},
\end{split}
\label{eq:kl}
\end{equation}
where $\overline{p}=\frac{1}{B}\sum\nolimits_{i=1}^{B}{p_i}\in{\mathbb R}^C$ represents the mean predicted probability distribution over a batch of target samples, and ${\omega}$ denotes a virtual label-distribution-aware weight vector.
In Eq.~(\ref{eq:kl}), the first term penalizes overconfidence in majority class predictions by leveraging the weight vector $\omega$, thereby encouraging predictions toward minority classes.
The second term promotes output diversity by penalizing low-entropy distributions—commonly referred to as diversity loss.
Notably, diversity loss is a special case of our formulation where $\omega$ is uniform, treating all categories equally.

Importantly, this loss operates in a fully unsupervised manner and does not require ground-truth annotations, making it particularly suitable for the target domain.
We define the explicit re-balancing loss $\mathcal{L}{RB}$ as:
\begin{equation}
    \mathcal{L}_{RB}(x^t, \omega) = D_{KL}(\overline{p}_{t} \| \omega),
    \label{eq:balance}
\end{equation}
where $\overline{p}{t}$ is the average prediction over the current batch of target samples $\{x^t_j\}_{j=1}^{B}$.

This re-balancing loss prevents the model from reinforcing incorrect category predictions in the target domain.
By introducing directionally diverse gradients, it encourages the classifier to better distinguish and recover underrepresented classes.
We restrict its application to the target domain, as applying it to the source domain—where labels are fixed—may lead to negative transfer.

\subsection{Overall Objective}
Since our method is specifically designed to improve worst-case performance, we retain the standard domain alignment component commonly used in existing UDA frameworks.
This design choice ensures that our method can be seamlessly integrated as a plug-and-play module into most existing UDA approaches.
The overall objective function of our proposed method is defined as follows:
\begin{equation}
    \mathcal{L}_{VILL} = \mathcal{L}_{RW}+\gamma\mathcal{L}_{DA}+\beta\mathcal{L}_{RB},
    \label{all loss}
\end{equation}
where $\gamma$, $\beta>0$ are hyperparameters that balance the domain alignment term and the re-balancing term, respectively.

\section{Experiment}
\textbf{Setup.}
To validate the effectiveness of our proposed method, we incorporate VILL with four UDA algorithms CDAN \cite{long2018conditional}, MDD \cite{zhang2019bridging}, ATDOC \cite{liang2021domain}, PDA \cite{bai2024prompt}, among which PDA is an adaptation method based on CLIP \cite{radford2021learning}, and the rest are methods proposed based on visual models.
We evaluate all methods on two commonly used image classification benchmarks, a popular four-domain dataset \textbf{OfficeHome} \cite{venkateswara2017deep} and a classic small-sized dataset \textbf{Office} \cite{saenko2010adapting}.
Please note that, in Office, due to the limited number of samples in the DSLR (D) domain, we focus our experiments on tasks where Amazon (A) and Webcam (W) serve as the target domains.

\textbf{Metric.}
Since there is no metric to measure category fairness in UDA tasks, we introduce Worst-$N$ accuracy, which is defined as the average accuracy of $N$ classes with the worst accuracy in the target domain.
We report Worst-5, Worst-10 accuracy, and the global accuracy in all benchmarks.

\textbf{Implementation details.}
VILL has two hyperparameters $\alpha$, $\beta$, we use $\alpha=5.0$ and $\beta = 0.05$ as default for all tasks. 
To demonstrate the versatility of VILL, we use the official code and hyperparameters for all baseline methods. 
We set the iteration number to 1000 in every training epoch.

\subsection{Comparison Results}
\begin{figure}[!t]
    \centering
    \begin{subfigure}[b]{0.49\linewidth}
        \centering
        \includegraphics[width=\linewidth]{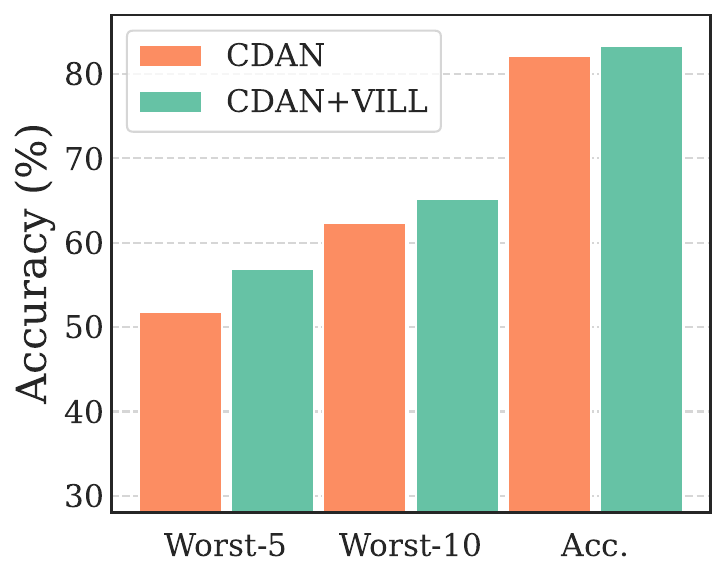}
        \caption{VILL deployed on CDAN}
        \label{fig:cdan-office}
    \end{subfigure}
    \hfill
    \begin{subfigure}[b]{0.49\linewidth}
        \centering
        \includegraphics[width=\linewidth]{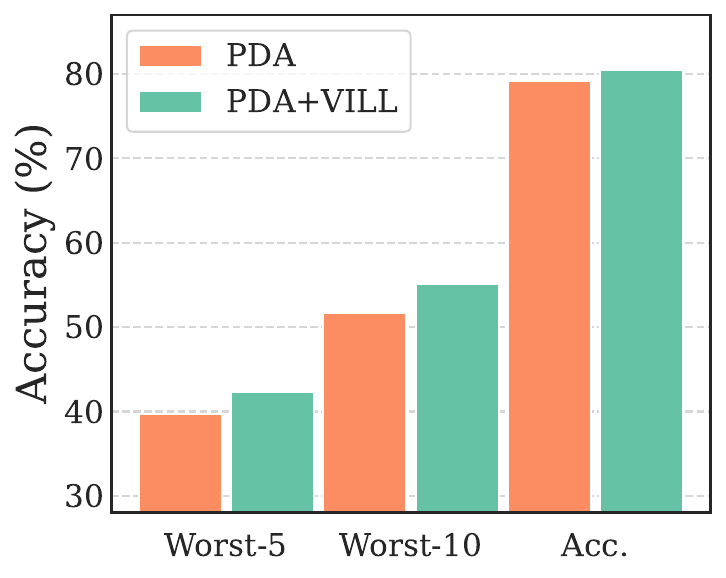}
        \caption{VILL deployed on PDA}
        \label{fig:pda-office}
    \end{subfigure}
    \caption{Worst-N accuracy and global accuracy of VILL on CDAN and PDA on Office dataset.}
    \label{fig:office}
    \vspace{-0.5cm}
\end{figure}

We compare the worst-case performance of VILL and UDA baseline methods in two datasets, as shown in Table~\ref{table:home-worst} and Figure~\ref{fig:office}.
These metrics highlight the performance on the most challenging classes, thus offering a clear view into category fairness.
It is obvious that all baseline methods suffer from severely poor category fairness, as the worst-N accuracy is far below the global one.
Across all four UDA baselines, integrating our proposed VILL module consistently improves the worst-case accuracy.
In terms of WAcc-5, we observe clear improvements across all methods.
For example, CDAN+VILL increases average worst-5 accuracy from 20.3\% to 26.8\% on OfficeHome while the global accuracy remains almost unchanged.

Since VILL has no requirements for the network structure, it can also be deployed in the CLIP-based UDA method.
For example, PDA+VILL(CLIP-ResNet-50) improves average worst-5 accuracy from 32.8\% to 35.1\%, which shows that VILL remains effective even with powerful pretrained models.
These results demonstrate that VILL enhances representation for poorly performing categories.

An important observation is that the average accuracy remains largely stable or slightly improves.
Especially, PDA+VILL achieves a slight increase in average accuracy from 74.5\% to 75.0\% in OfficeHome.
As shown in Fig.~\ref{fig:cdan-office}, \ref{fig:pda-office}, VILL also improves the global accuracy of the two UDA methods on the Office dataset while enhancing their category fairness.
It confirms that VILL does not trade off overall performance for fairness; it improves both worst-case and global performance in some scenarios.

\subsection{Ablation study}
We conduct an ablation study of VILL on the OfficeHome dataset using CDAN as the base UDA method.
The results are summarized in Table~\ref{tab:ablation}.
The baseline CDAN model yields the lowest worst-case performance, reflecting poor category fairness.
Introducing the re-weighting mechanism improves both Worst-5 and Worst-10 metrics.
Similarly, employing the re-balancing strategy leads to a more substantial gain in worst-case performance.
Finally, combining both modules yields the best worst-case performance (26.8\% on Worst-5 and 34.5\% on Worst-10), which validates the effectiveness of both modules inside VILL in improving category fairness.

\setlength{\tabcolsep}{2.0pt}
    \begin{table}[!t]
    \caption{\textbf{Ablation study.} Worst-case performance and accuracy (\%) on OfficeHome dataset of VILL on CDAN method.}
    \label{tab:ablation}
        \centering
        \vspace{-3mm}
        \centering
        \resizebox{0.9\linewidth}{!}
        {
            \begin{tabular}{cc|ccc}
            \toprule
            {Re-weighting} & {Re-balancing} & {Worst-5} & {Worst-10} & {Accuracy} \\
            \midrule
            \XSolidBrush & \XSolidBrush & 20.3 & 28.7 & 68.0 \\
            \CheckmarkBold & \XSolidBrush & 22.4 & 31.2 & 68.7 \\
            \XSolidBrush & \CheckmarkBold & 26.0 & 33.7 & 68.0 \\
            \CheckmarkBold & \CheckmarkBold & 26.8 & 34.5 & 67.9 \\
            \bottomrule
            \end{tabular}
        }
    
    \end{table}

\section{Conclusion}
In this paper, we investigated the issue of category fairness in Unsupervised Domain Adaptation (UDA) and introduced a new evaluation metric to better assess the performance disparities across categories.
Our empirical analysis reveals that many existing UDA methods exhibit severe imbalances, and we propose VILL to alleviate this phenomenon.
VILL consists of a re-weighting mechanism to emphasize difficult categories during source training, and a re-balancing strategy that adjusts the decision boundaries.
Extensive experiments demonstrate that VILL substantially improves worst-case category performance while maintaining overall accuracy.
We believe our work offers meaningful insights toward building more reliable transfer learning systems.

% -------------------------------------------------------------------------

\section{Acknowledgements}
This work was supported by the National Natural Science Foundation of China under Grants (62276256, U2441251).

\section{Ethical Standards Statement}
This is a numerical simulation study for which no ethical approval was required.

\bibliographystyle{IEEEbib}
\bibliography{main}

@string(MLJ = "Machine Learning")

@string(ICML = "Proc. ICML")

@string(CVPR = "Proc. CVPR")

@string(ICCV = "Proc. ICCV")

@string(ECCV = "Proc. ECCV")

@string(NeurIPS = "Proc. NeurIPS")

@string(AAAI = "Proc. AAAI")

@string(ICLR = "Proc. ICLR")

@article{ben2010theory,
  title={A theory of learning from different domains},
  author={Ben-David, Shai and Blitzer, John and Crammer, Koby and Kulesza, Alex and Pereira, Fernando and Vaughan, Jennifer Wortman},
  journal=MLJ,
  volume={79},
  number={1},
  pages={151--175},
  year={2010},
  publisher={Springer}
}

@inproceedings{liang2021domain,
  title={Domain adaptation with auxiliary target domain-oriented classifier},
  author={Liang, Jian and Hu, Dapeng and Feng, Jiashi},
  booktitle=CVPR,
  year={2021}
}

@inproceedings{long2018conditional,
  title={Conditional adversarial domain adaptation},
  author={Long, Mingsheng and Cao, Zhangjie and Wang, Jianmin and Jordan, Michael I},
  booktitle=NeurIPS,
  year={2018}
}

@inproceedings{zhang2019bridging,
  title={Bridging theory and algorithm for domain adaptation},
  author={Zhang, Yuchen and Liu, Tianle and Long, Mingsheng and Jordan, Michael},
  booktitle=ICML,
  year={2019}
}

@inproceedings{saenko2010adapting,
  title={Adapting visual category models to new domains},
  author={Saenko, Kate and Kulis, Brian and Fritz, Mario and Darrell, Trevor},
  booktitle=ECCV,
  year={2010}
}

@inproceedings{venkateswara2017deep,
  title={Deep hashing network for unsupervised domain adaptation},
  author={Venkateswara, Hemanth and Eusebio, Jose and Chakraborty, Shayok and Panchanathan, Sethuraman},
  booktitle=CVPR,
  year={2017}
}

@article{ganin2016domain,
  title={Domain-adversarial training of neural networks},
  author={Ganin, Yaroslav and Ustinova, Evgeniya and Ajakan, Hana and Germain, Pascal and Larochelle, Hugo and Laviolette, Fran{\c{c}}ois and Marchand, Mario and Lempitsky, Victor},
  journal={JMLR},
  volume={17},
  number={1},
  pages={2096--2030},
  year={2016},
  publisher={JMLR. org}
}

@inproceedings{tzeng2017adversarial,
  title={Adversarial discriminative domain adaptation},
  author={Tzeng, Eric and Hoffman, Judy and Saenko, Kate and Darrell, Trevor},
  booktitle=CVPR,
  year={2017}
}

@inproceedings{cui2019class,
  title={Class-balanced loss based on effective number of samples},
  author={Cui, Yin and Jia, Menglin and Lin, Tsung-Yi and Song, Yang and Belongie, Serge},
  booktitle=CVPR,
  year={2019}
}

@inproceedings{zou2018unsupervised,
  title={Unsupervised domain adaptation for semantic segmentation via class-balanced self-training},
  author={Zou, Yang and Yu, Zhiding and Kumar, BVK and Wang, Jinsong},
  booktitle=ECCV,
  year={2018}
}

@inproceedings{he2016deep,
  title={Deep residual learning for image recognition},
  author={He, Kaiming and Zhang, Xiangyu and Ren, Shaoqing and Sun, Jian},
  booktitle=CVPR,
  year={2016}
}

@inproceedings{vu2019advent,
  title={Advent: Adversarial entropy minimization for domain adaptation in semantic segmentation},
  author={Vu, Tuan-Hung and Jain, Himalaya and Bucher, Maxime and Cord, Matthieu and P{\'e}rez, Patrick},
  booktitle=CVPR,
  year={2019}
}

@inproceedings{ge2019mutual,
  title={Mutual Mean-Teaching: Pseudo Label Refinery for Unsupervised Domain Adaptation on Person Re-identification},
  author={Ge, Yixiao and Chen, Dapeng and Li, Hongsheng},
  booktitle=ICLR,
  year={2019}
}

@inproceedings{kim2019self,
  title={Self-training and adversarial background regularization for unsupervised domain adaptive one-stage object detection},
  author={Kim, Seunghyeon and Choi, Jaehoon and Kim, Taekyung and Kim, Changick},
  booktitle=ICCV,
  year={2019}
}

@inproceedings{vs2021mega,
  title={Mega-cda: Memory guided attention for category-aware unsupervised domain adaptive object detection},
  author={Vs, Vibashan and Gupta, Vikram and Oza, Poojan and Sindagi, Vishwanath A and Patel, Vishal M},
  booktitle=CVPR,
  year={2021}
}

@inproceedings{dwork2012fairness,
  title={Fairness through awareness},
  author={Dwork, Cynthia and Hardt, Moritz and Pitassi, Toniann and Reingold, Omer and Zemel, Richard},
  booktitle={Proc. ITCS},
  //pages={214--226},
  year={2012}
}

@inproceedings{bai2024prompt,
  title={Prompt-based distribution alignment for unsupervised domain adaptation},
  author={Bai, Shuanghao and Zhang, Min and Zhou, Wanqi and Huang, Siteng and Luan, Zhirong and Wang, Donglin and Chen, Badong},
  booktitle={Proc. AAAI},
  //pages={729--737},
  year={2024}
}

@inproceedings{radford2021learning,
  title={Learning transferable visual models from natural language supervision},
  author={Radford, Alec and Kim, Jong Wook and Hallacy, Chris and Ramesh, Aditya and Goh, Gabriel and Agarwal, Sandhini and Sastry, Girish and Askell, Amanda and Mishkin, Pamela and Clark, Jack and others},
  booktitle={Proc. ICML},
  //pages={8748--8763},
  year={2021},
}

@book{quinonero2022dataset,
  title={Dataset shift in machine learning},
  author={Qui{\~n}onero-Candela, Joaquin and Sugiyama, Masashi and Schwaighofer, Anton and Lawrence, Neil D},
  year={2022},
  publisher={Mit Press}
}

@inproceedings{cicek2019unsupervised,
  title={Unsupervised domain adaptation via regularized conditional alignment},
  author={Cicek, Safa and Soatto, Stefano},
  booktitle={Proc. ICCV},
  //pages={1416--1425},
  year={2019}
}

@inproceedings{chang2019domain,
  title={Domain-specific batch normalization for unsupervised domain adaptation},
  author={Chang, Woong-Gi and You, Tackgeun and Seo, Seonguk and Kwak, Suha and Han, Bohyung},
  booktitle={Proc. CVPR},
  //pages={7354--7362},
  year={2019}
}

@article{singhal2023domain,
  title={Domain adaptation: challenges, methods, datasets, and applications},
  author={Singhal, Peeyush and Walambe, Rahee and Ramanna, Sheela and Kotecha, Ketan},
  journal={IEEE Access},
  volume={11},
  pages={6973--7020},
  year={2023},
  publisher={IEEE}
}

\end{document}